\documentclass[10pt]{article}

\usepackage{times}
\usepackage{graphicx}
\usepackage{epsfig}
\usepackage{amsmath}
\usepackage{amssymb}
\usepackage{booktabs}
\usepackage[table,xcdraw]{xcolor}
\usepackage{algorithm}
\usepackage{algorithmic}
\usepackage{multirow}
\usepackage{subcaption}
\usepackage{hyperref}

\newcommand{\ej} {\color{black}}

\newcommand{\eg}{{\em e.g.}}           
           
\newcommand{\ie}{{\em i.e.}}

\begin{document}

\title{HomographyAD: Deep Anomaly Detection Using Self Homography Learning}

\author{
Jongyub Seok$^{1,2}$\thanks{Equal contribution.} \quad Chanjin Kang$^{1,3}$\footnotemark[1] \\
$^1$AIRS Company, Hyundai Motor Group, Seoul, Korea \\
$^2$Currently with LG Energy Solution \\
$^3$Currently with Dfinite Inc. \\
{\tt\small ffighting.seok@gmail.com} \quad {\tt\small cjkang@dfinite.ai}
}

\date{}
\maketitle

\begin{abstract}
{\ej
   Anomaly detection (AD) is a task that distinguishes normal and abnormal data, which is important for applying automation technologies of the manufacturing facilities. For MVTec dataset that is a representative AD dataset for industrial environment, many recent works have shown remarkable performances. However, the existing anomaly detection works have a limitation of showing good performance for fully-aligned datasets only, unlike real-world industrial environments. To solve this limitation, we propose HomographyAD, a novel deep anomaly detection methodology based on the ImageNet-pretrained network, which is specially designed for actual industrial dataset. Specifically, we first suggest input foreground alignment using the deep homography estimation method. In addition, we fine-tune the model by self homography learning to learn additional shape information from normal samples. Finally, we conduct anomaly detection based on the measure of how far the feature of test sample is from the distribution of the extracted normal features. By applying our proposed method to various existing AD approaches, we show performance enhancement through extensive experiments.}

\end{abstract}
\section{Introduction}
{\ej Recently, as computer vision technology develops, manufacturing plants have attempted applying automation technologies to reduce costs \cite{babic2021image}. Specifically, in the quality inspection automation system, it is important to distinguish whether the current state of a manufacturing facility is normal or abnormal, \ie, \emph{anomaly detection} \cite{pang2021deep}. With the recent development of deep learning, many anomaly detection studies have been proposed, showing remarkable performances for various public industrial datasets \cite{huang2020surface, mishra2021vt}, \eg, MVTec dataset \cite{bergmann2019mvtec}. In this work, we focus on developing the anomaly detection method that is specially designed for real-world industrial environment.}


\begin{figure}[t]
    \centering
    \includegraphics[width=230pt]{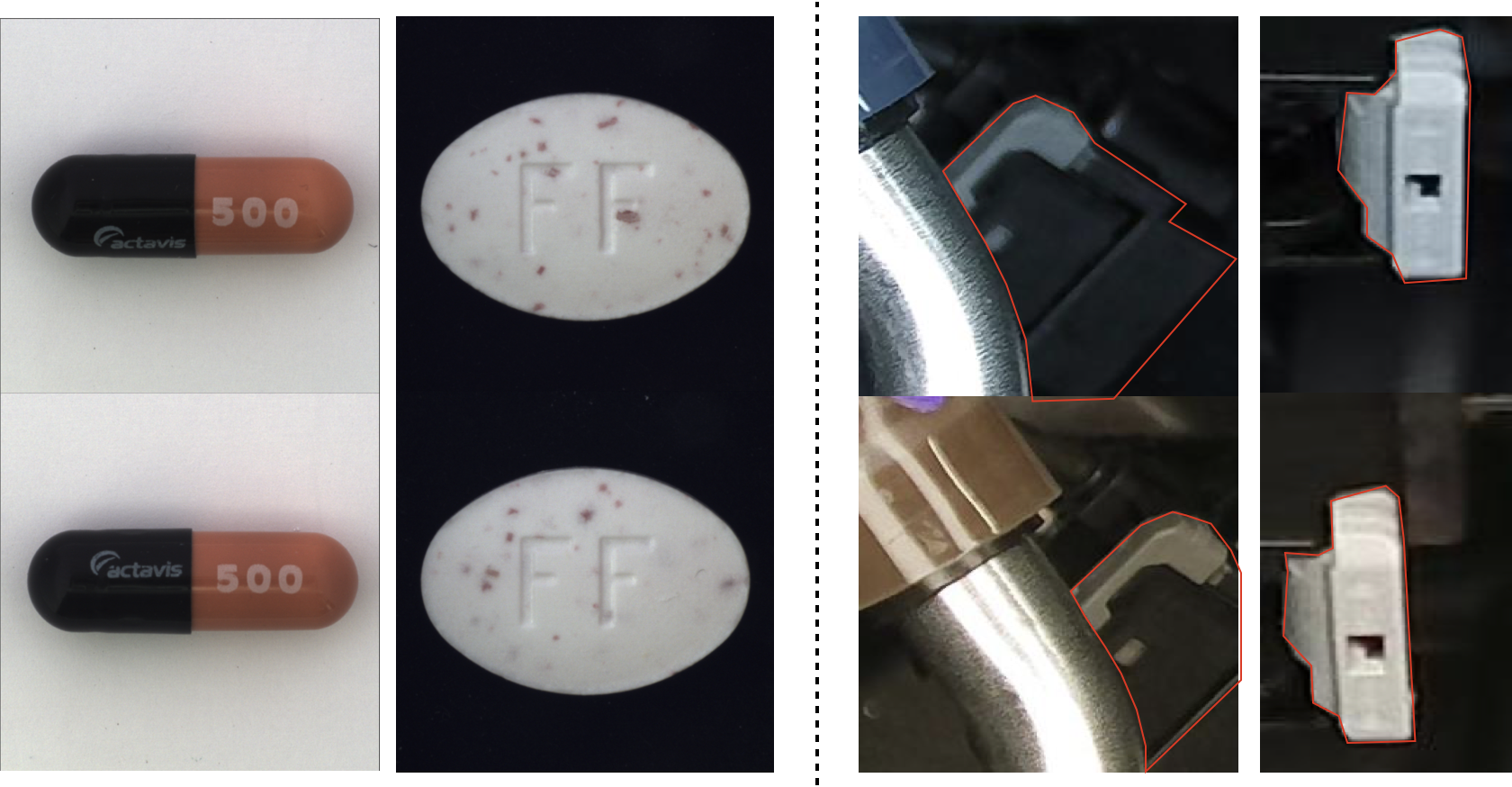}
    \caption{{\ej Examples of (Left) public MVTec dataset and (Right) real-world industrial dataset. The images in each column are normal images of the same class. The red outline in the right two columns represents the foreground, the target of the anomaly detection, and the rest represent the background.}
}
   \label{fig:fig1}
\end{figure}

{\ej For actual industrial environment, we deal with the samples of anomaly detection that have the following characteristics: First, they focus on specific objects for elaborate inspection purposes, resulting in sophisticated yet recurrent patterns as shown in Figure \ref{fig:fig1}, \ie, low intra-class variance and high inter-class variance. In addition, there are few abnormal samples to be learned, compared to many normal samples in the industrial environment. Due to these characteristics, many works have developed unsupervised anomaly detection methods that use only normal samples for training the model and distinguish between normal and abnormal samples in the evaluation phase \cite{ruff2018deep, schlegl2017unsupervised, akcay2018ganomaly}.
}


{\ej
To this end, more recently, instead of training the model on normal data separately, some works \cite{bergman2020deep, cohen2020sub, roth2022towards, rippel2021modeling, defard2021padim} have attempted to leverage a pre-trained network on a large dataset such as ImageNet \cite{deng2009imagenet}, denoted by a \emph{pre-trained network-based anomaly detection method (PAD)} in this paper. They assumed that since the pre-trained network already learns various features from large samples, we are able to distinguish normal and abnormal samples based on these useful features. Furthermore, most feature maps at the same location contain same object information, and the degree of normal is measured based on how far it has deviated from the distribution of information contained in each location of the feature map. These methods not only showed significant performance on the MVTec dataset that is a representative industrial dataset, but also did not spend time training on normal data, making them convenient to apply in real industrial environments. Thus, in this work, we also exploit the pre-trained network as our feature extractor for effective anomaly detection.
}


{\ej However, the existing PAD methods suffer from poor performances when applied to the real-world automated anomaly detection system, unlike the public MVTec dataset. Figure \ref{fig:fig1} represents the difference between the public MVTec dataset (left) and the real-world industrial dataset (right). For example, both the capsules and pills from the MVTec dataset are located in the center of the image and the angles are equally aligned. On the other hand, the connector parts from the actual industrial dataset we handled, marked with a red outline, are not well aligned. That is, it is noteworthy that the foreground part of the MVTec dataset aligns well, whereas that of the actual industrial environment is not located in the same location on the image. This mismatch of alignment in the real-world dataset challenges to the assumption of the previous PAD approaches that the same location of the feature map will contain the same object information, resulting into performance degradation.}


{\ej Accordingly, we assume that the input foreground alignment plays a key role in anomaly detection, and better alignment leads to improved performance on real-world industrial dataset. To verify our assumption, we generated the synthetic MVTec datasets by modifying the degree of alignment and investigated the effect of different degrees of alignment (See Section \ref{subsec:datasets}.). Here, we utilized deep homography estimation method \cite{detone2016deep}, which has been widely used in 2D transformation mapping between two images, to match the foreground alignment in the dataset. As a result, interestingly, we observed that the better foreground alignment contributed to the better performances for anomaly detection, leading to improved performance for various PAD approaches on different datasets (See Table \ref{tbl:tbl1}).

In this paper, we propose HomographyAD, an anomaly detection framework built with the ImageNet pre-trained network, based on deep homography estimation, for real-world industrial environment. First, we suggest \emph{input-level} foreground alignment by deep homography estimation. Then, in addition to the effect of the input-level from homography estimation, we also explore that of the \emph{feature-level} on the pre-trained network. To this end, we further fine-tune the pre-trained network on normal samples by 2D transformation mapping to learn additional features, including shape information, \ie, \emph{self homography learning}. That is, our proxy task of self-supervised learning (SSL) is to learn its own degree of alignment using an aligned dataset. Finally, we measure how far the feature of test sample is from the distribution of the extracted normal features as anomaly score, which is followed by the conventional PAD approaches \cite{cohen2020sub, roth2022towards, rippel2021modeling, defard2021padim, bergman2020deep}. In summary, our contributions are as follows:
\begin{itemize}
    \item We demonstrate that the performance degradation of existing PAD methods on the real-world industrial datasets can be handled by input foreground alignment.
    
    \item We propose a method to effectively learn the characteristics of normal data by self homography learning as our SSL task.
    
    \item By applying our proposed method to various existing PAD approaches, we show performance enhancement through extensive experiments.
\end{itemize}

}

\section{Related work}

\subsection{Industrial Anomaly Detection}

Early in the industrial AD, some works \cite{schlegl2017unsupervised, akcay2018ganomaly, bergmann2019mvtec} have tried to train reconstruction of normal images. These works assumed that if model trained to reconstruct normal images, the normal part will be reconstructed well but the anomal part will not at test stage.
Schleg \textit{et al.} \cite{schlegl2017unsupervised} and Akcay \textit{et al.} \cite{akcay2018ganomaly} attempted AD using the Generative Adversarial Network \cite{goodfellow2014generative}.
MVTec-AD \cite{bergmann2019mvtec} attempted AD using AutoEncoder.

Golan \textit{et al.} \cite{golan2018deep} proposed an AD model using self-supervised learning.
Hendrycks \textit{et al.} \cite{hendrycks2019using} trained not only rotation but also translation.
Sohn \textit{et al.} \cite{sohn2020learning} showed that high AD performance can be achieved by using simple classifiers such as Kernel density estimation (KDE) with features trained by self supervised learning.

Recently, anomaly detection methods using pretrained networks (PAD) have been proposed.
SPADE \cite{cohen2020sub} divided the image into patches and measured the anomal score by euclid distance for each patch.
Patchcore \cite{roth2022towards} uses a memory bank that collects neighborhood aware features.
Rippel \textit{et al.} \cite{rippel2021modeling} and Defard \textit{et al.} \cite{defard2021padim} tried to use mahalanobis distance \cite{mahalanobis1936generalized} to calculate distance between features.
Unlike other methods, these PAD methods have a strong advantage that they do not require separate resources for training.
This advantage is an important point that can be easily applied in manufacturing plant environments.
Therefore, in this work we focused on improving the performance of PAD methods through minimum additional training process.

\subsection{Deep Homography Estimation}

Homography refers to 2D transformation mapping between two images lying on the same plane.
Thus, homography estimation (HE) refers to a method of finding a homography matrix between these two images.
The existing homography estimation mainly used the features like SIFT \cite{lowe2004distinctive}, ORB \cite{rublee2011orb} to find interesting points in two images.
And then match corresponding points of each features of them.
Commomly, RANSAC is used to avoid incorrect matching.
However, this methods has poor performance due to its sensitivity to external conditions like illuminance.

Recently, this limitation is handled by applying the deep learning method to homography estimation, which is called deep homography estimation (DHE).
The first DHE model was \cite{detone2016deep}.
They used a transformed image created by applying random perturbation to four corners of the image as training data.
They showed better performance than the existing homography estimation method without the labeling cost.
Erlik \textit{et al.} \cite{erlik2017homography} learned hierarchical features in a supervised learning manner.
CLKN \cite{chang2017clkn} incorporated the LukasKanade \cite{baker2004lucas} method into the deep learning framework.
Zeng \textit{et al.} \cite{zeng2018rethinking} applied a model that outputs the flow for each pixel rather than the pixel value.
Nguyen \textit{et al.} \cite{nguyen2018unsupervised} performed the unsupervised deep homography estimation.
Zhang \textit{et al.} \cite{zhang2020content} used a mask predictor network, focusing only on foreground.
Koguciuk \textit{et al.} \cite{koguciuk2021perceptual} used unsupervised bi-directional triplet loss and homography matrix regression loss.

In this work, we employ the DHE model based on \cite{koguciuk2021perceptual}
for input foreground alignment and self homography learning as our proxy task for self-supervised learning.

\section{Proposed Method}
{\ej In this section, we describe the overall framework of our HomographyAD for real-world industrial environment. Based on the ImageNet pre-trained network as a backbone feature extractor, our framework mainly comprises the following three steps: (i) aligning input foreground by deep homography estimation method, (ii) fine-tuning the backbone network through self homography learning to learn additional shape information useful for anomaly detection, and (iii) anomaly detection based on the distance of feature from normal features' distribution.

} 




\subsection{Input Foreground Alignment}
\label{subsec:input foreground alignment}
{\ej For input foreground alignment, we employ the deep homography estimation (DHE) method. Homography estimation aims to find 2D transformation that can be mapped to different images that exist on the same plane. To this end, we randomly select a template image from the normal data for reference. Then, we train the DHE model by predicting the changed values $(\Delta x_{i},\Delta y_{i})$ for each coordinate $(x_{i},y_{i})$ of four corners of an image, where in this work, \cite{koguciuk2021perceptual} is chosen for our base model. The predictions for eight delta values are then converted into homography matrix using DLT \cite{page2005multiple}. Finally, based on the learned model, we infer the homography matrix 
of target samples by the template image and align them based on the resulting homography matrix.

However, since the model only learns the homography matrix for its own transformed image, it is difficult to obtain the homography matrix of target samples that are different from the template image. To handle this issue, we train the model using augmentation to learn the various types of images. In this work, the augmentation includes hue, bright, pepper, and salt. 
}



\begin{figure}[tp!]
    \centering
    \includegraphics[width=260pt]{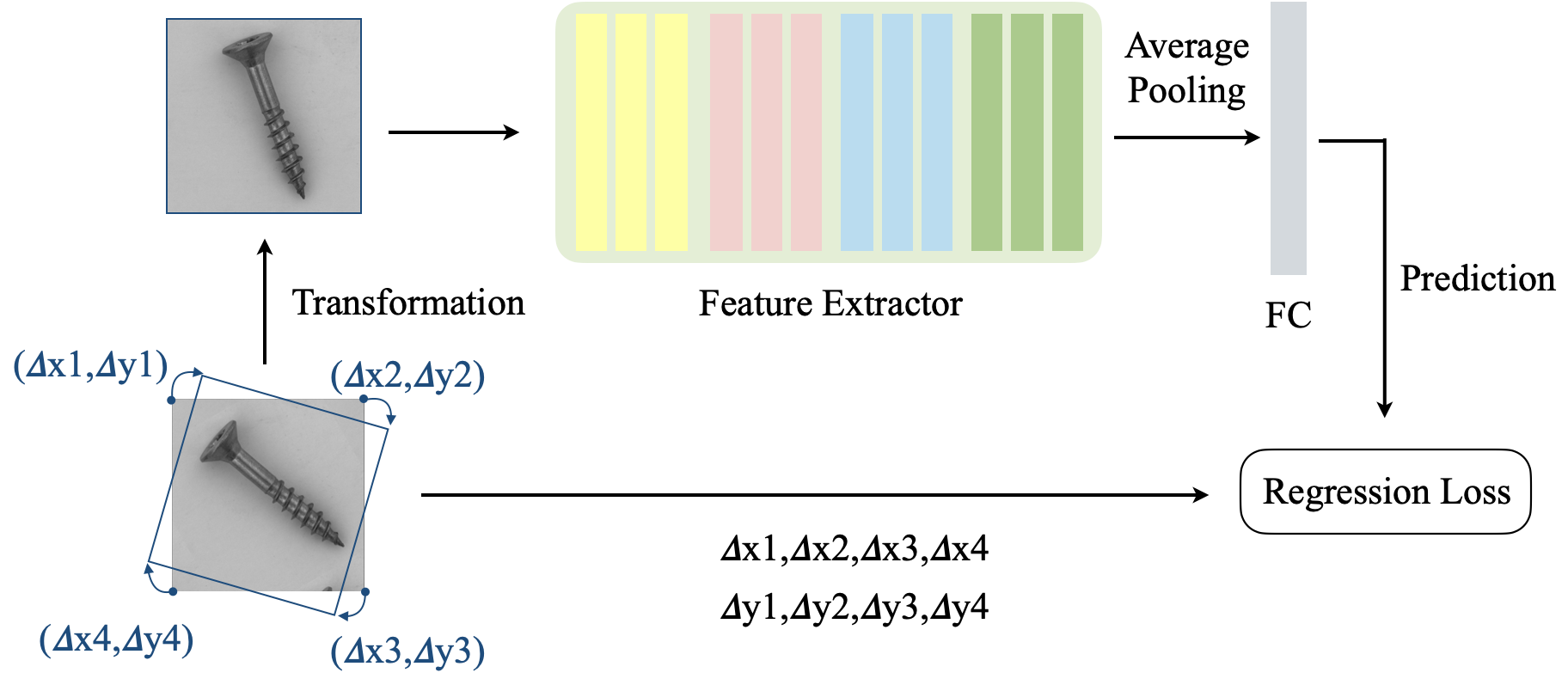}
    \caption{Illustration of self homography learning process.}
    \label{fig:fig3}
\end{figure}

\begin{figure}[tp!]
\centering
\includegraphics[width=230pt]{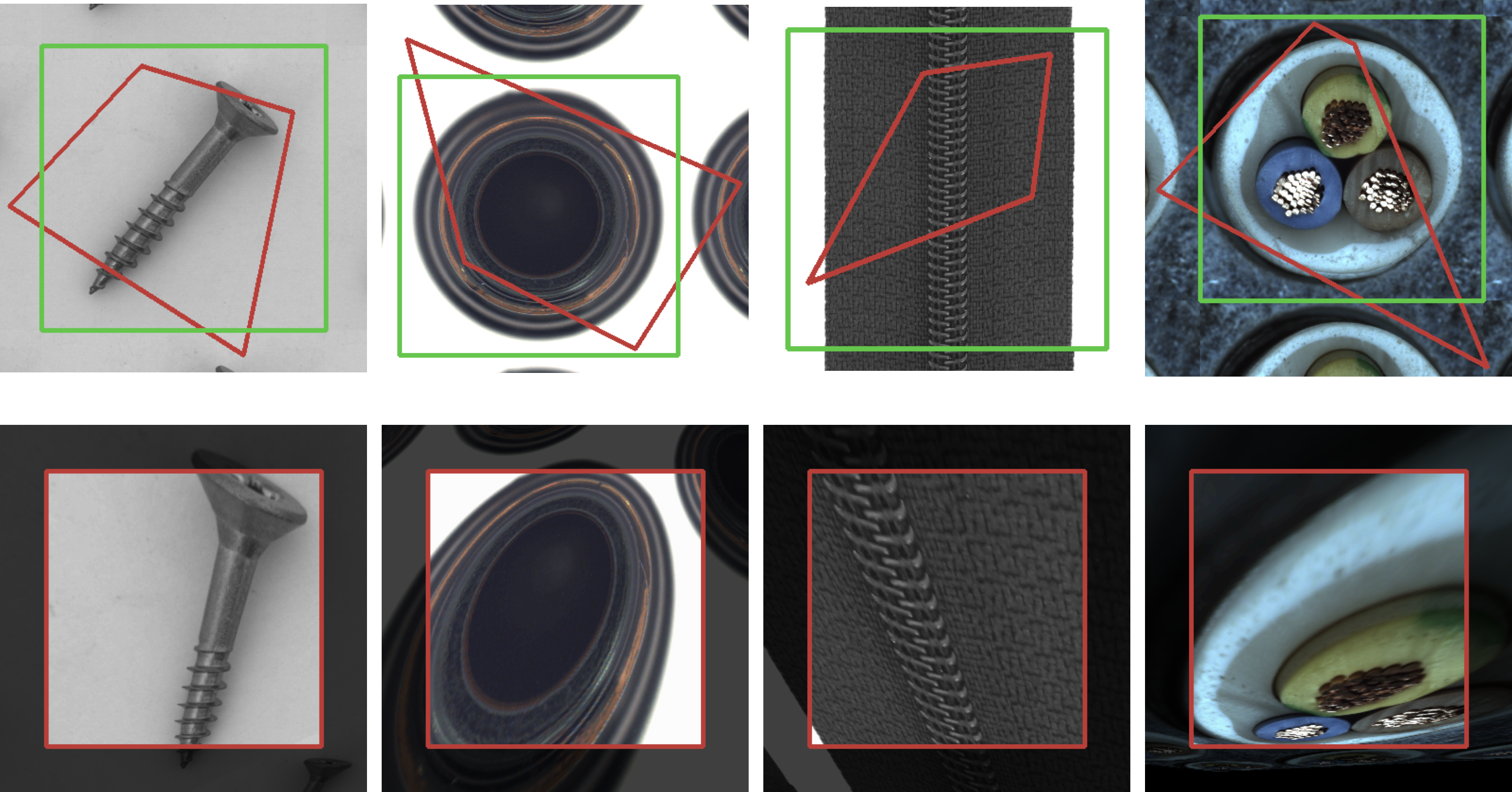}
\caption{{\ej Visualization of images from the MVtec dataset (1st row) and their transformed images (2nd row). Images of 2nd row are generated such that the images of 1st row are transformed from red outline to green one.}}
\label{fig:fig2}
\end{figure}

\begin{figure*}[thp!]
     \centering
     \begin{subfigure}[b]{0.37\textwidth}
         \centering
         \includegraphics[width=\textwidth]{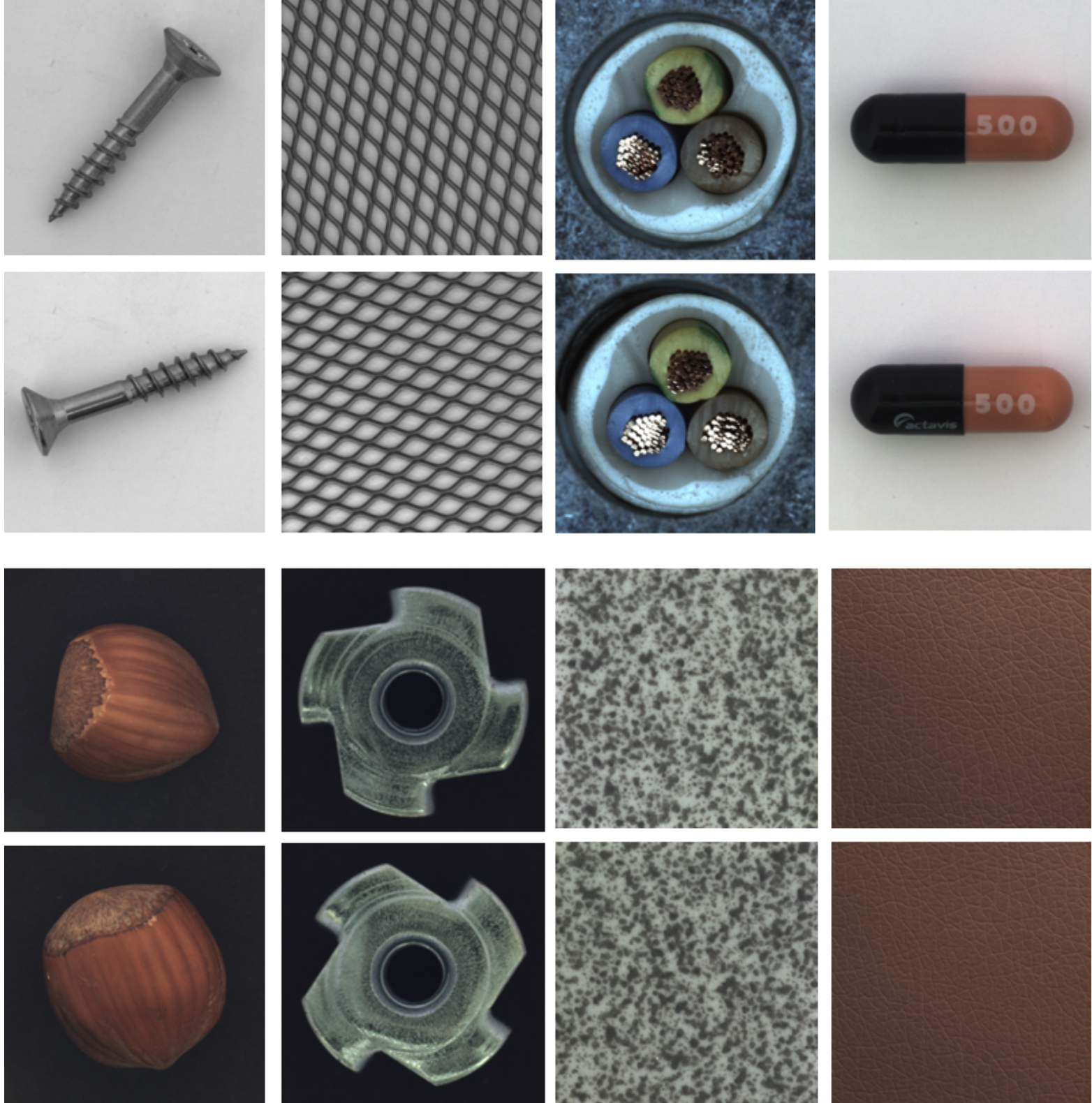}
         \caption{}
      \label{fig:fig4}
     \end{subfigure}
     \hfill
     \begin{subfigure}[b]{0.185\textwidth}
        \centering
        \includegraphics[width=\textwidth]{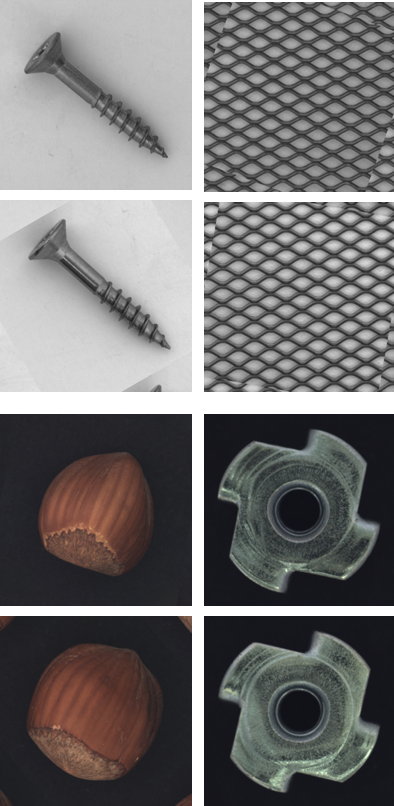}
        \caption{}
   \label{fig:fig5}
     \end{subfigure}
     \hfill
     \begin{subfigure}[b]{0.376\textwidth}
        \centering
        \includegraphics[width=\textwidth]{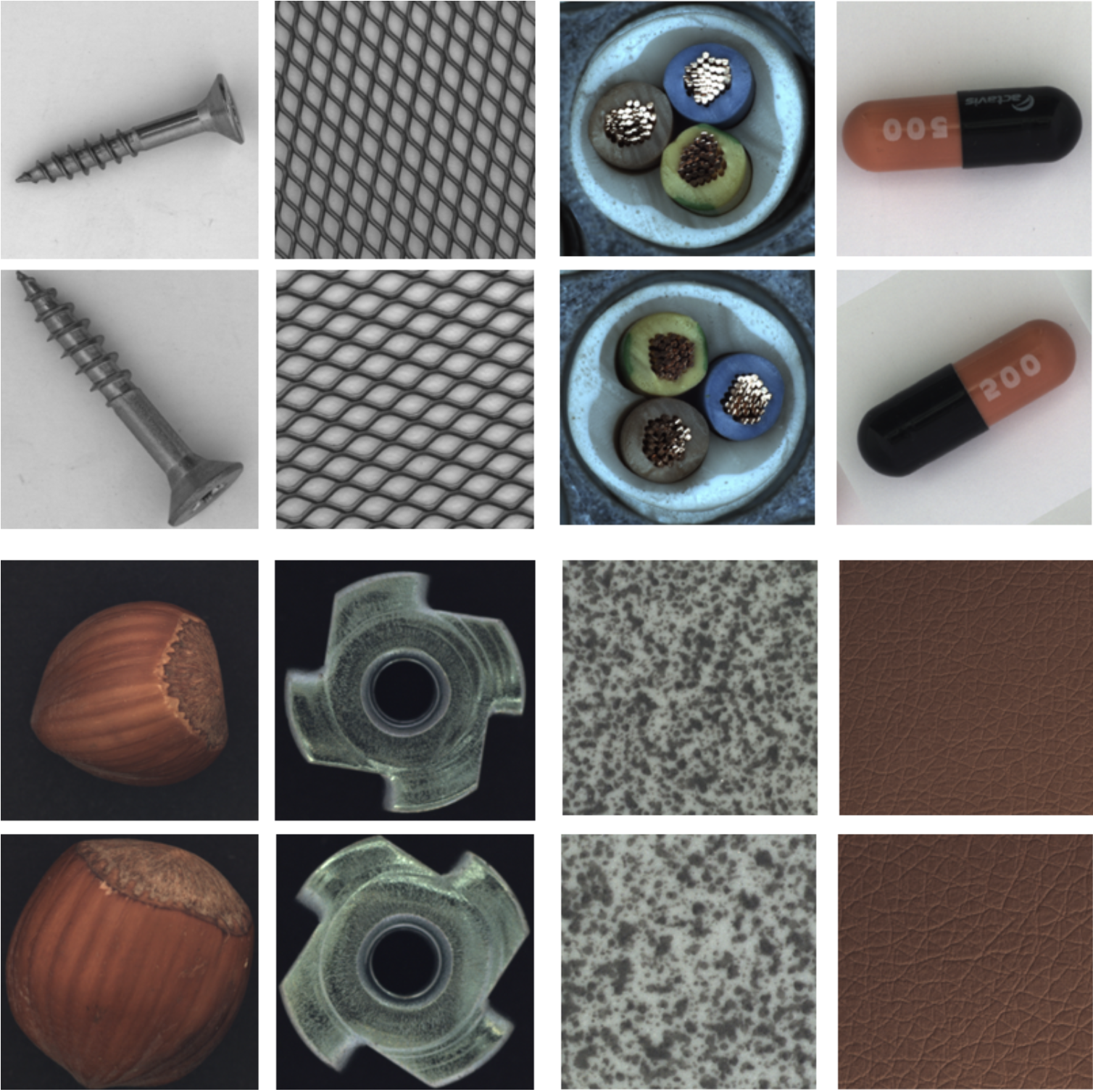}
        \caption{}
      \label{fig:fig6}
     \end{subfigure}
    \caption{{\ej Examples of the MVTec dataset. (a) Image samples of original MVTec dataset. Screw, grid, cable, capsule, hazelnut, metalnut, tile, and leather classes in order from top left to bottom right. Left four classes are not aligned (\emph{Misaligned Classes of Original MVTec}). Right fours classes are aligned  (\emph{Aligned Classes of Original MVTec}) (b) Image samples of Synthetic MVTec w/ Alignment. Compared to Figure \ref{fig:fig4}, alignment of first and second row images of each class fits. (c) Image samples of Synthetic MVTec w/o Alignment. Screw, grid, cable, capsule, hazelnut, metalnut, tile, and leather classes in order from top left to bottom right. They aligned worse than Figure \ref{fig:fig4}.}}
    \label{fig:mvtec_dataset}
\end{figure*}

\subsection{Self Homography Learning}
{\ej It is known that the ImageNet pre-trained network focuses on the textures of image \cite{geirhos2018imagenet}. This means that ImageNet pre-trained feature is rich in texture information, but shape information is insufficient. Therefore, we can infer that PAD is vulnerable to a class that has various shape changes. This fact implies that by injecting shape information into a pre-trained network, we can improve AD performance.

Meanwhile, many works have proposed self-supervised learning methods using image transformation such as classification of rotation angle \cite{gidaris2018unsupervised} and translation \cite{sohn2020learning}. In this work, beyond the rotation angle classification, it is extended to a regression task, which is inspired by 2D homography estimation. As a result, we obtain the following advantages: First, model can learn richer features than classification because the regression can induce a model to represent a more continuous range of degrees. Second, shape information of normal data can be additionally learned because the model focuses on shape information rather than texture information during the transformation. Thus, we assume that if we fine-tune the ImageNet pre-trained network to match the 2D general transformation from the normal images, we could obtain a feature extractor containing both texture information from ImageNet and shape information from the AD dataset.


Accordingly, we propose self-homography learning to learn 2D general transformation from normal data. Figure \ref{fig:fig3} represents the overall process of the self homography learning method. First, the model receives an image to which random 2D perturbation (transformation) is applied as an input. Here, it is assumed that input image is well-aligned from the previous step described in Section \ref{subsec:input foreground alignment}. The difference from random perturbation to the four corners, expressed as $(\Delta x_{i},\Delta y_{i})$, becomes the ground truth model learn. Passing through the pre-trained network with modified fully connected layer, the model predicts a total of four outputs, $(\Delta x_{i}',\Delta y_{i}')$. Finally, we can calculate regression loss by using the predictions and the ground truth values, as follows:
\begin{equation} 
    L = \sum_{i=1}^{4}\left| \Delta _{i}-\Delta' _{i} \right|^{2},
\end{equation}
where $\Delta _{i}$ denotes an ordered pair $(\Delta x_{i},\Delta y_{i})$. As a result, we can obtain a fine-tuned neural network that has additional shape information.

As shown in Figure \ref{fig:fig2}, we visualize images from the MVtec dataset (1st row) and their transformed images (2nd row). By random 2D perturbation, images of 2nd row are generated such that the images of 1st row are transformed from red outline to green one. To know which transformation is applied, the model has to focus on angle changes of screw, distortion of bottle shape, changes of zipper line, distortion of cable shape. That is, it can be seen that the texture information is not considered important in transformation.
}

\subsection{Homography-Guided Anomaly Detection}
{\ej By using the fine-tuned network obtained by our self homography learning method, we conduct anomaly detection task. Following the existing PAD approaches \cite{cohen2020sub, roth2022towards, rippel2021modeling, defard2021padim}, we extract features of an input image and measure the anomaly score based on how far the test feature is from the distribution of the extracted normal features. Since we only replace the ImageNet pre-trained network with a self homography fine-tuned network while using the existing PAD method, our self homography learning can be applied to all PAD methods.
}

\section{Experiments}
In this section, we conducted a quantitative and qualitative evaluation of our proposed method.
To demonstrate the superiority of our method, we experiment to obtain answers to the questions below.

Q1 : Does the PAD method performance decline as the alignment of dataset gets worse?
(See Section \ref{subsec:effect of align}.)

Q2: Does the PAD method using the fine-tuned network by self homography learning perform better than the one using ImageNet pre-trained network? (See Section \ref{subsec:effect of SHL}.)

\begin{table*}[thp!]
\centering
\setlength{\tabcolsep}{20pt}
\renewcommand{\arraystretch}{1.1}
\caption{Performance comparison of the state-of-the-art studies between different alignment condition datasets.}
\label{tbl:tbl1}
\begin{center}
\resizebox{\textwidth}{!}{%
\begin{tabular}{ll|c|c|c}
\cline{2-5}
 &
  \textbf{Method} &
  \begin{tabular}[c]{@{}c@{}}\textbf{Synthetic MVTec} \\ \textbf{w/o Alignment}\end{tabular} &
  \begin{tabular}[c]{@{}c@{}}\textbf{Misaligned Class of}\\\textbf{Original MVTec}\end{tabular} &
  \begin{tabular}[c]{@{}c@{}}\textbf{Synthetic MVTec} \\ \textbf{w/ Alignment}\end{tabular} \\ \cline{2-5} 
\multirow{4}{*}{\rotatebox{90}{Image}}   
                       & PaDiM \cite{defard2021padim}    & 89.69 (-3.26) & 92.94 & 94.85 (\textbf{+1.91}) \\
                       & PatchCore \cite{roth2022towards} & 94.83 (-2.81) & 97.64 & 96.85 (-0.79) \\
                       & SPADE \cite{cohen2020sub}        & 64.77 (-3.03) & 67.8  & 72.33 (\textbf{+4.53}) \\
                       & Mah.AD \cite{rippel2021modeling} & 73.75 (-6.41) & 80.16 & 80.36 (\textbf{+0.20}) \\ \cline{2-5} 
\multirow{3}{*}{\rotatebox{90}{Pixel}} 
                       & PaDiM \cite{defard2021padim}    & 95.88 (-1.66) & 97.54 & 97.84 (\textbf{+0.30}) \\
                       & PatchCore \cite{roth2022towards} & 96.90 (-0.78) & 97.67 & 97.39 (-0.29) \\
                       & SPADE \cite{cohen2020sub}        & 91.36 (-2.36) & 93.71 & 92.89 (-0.83) \\ \cline{2-5} 
\end{tabular}%
}
\end{center}
\end{table*}

\subsection{Datasets}
\label{subsec:datasets}
\subsubsection{Original MVTec Dataset}

MVTec is representative AD dataset of industrial environments.
It consists of total 15 classes.
Train data is composed of only normal data, and test data is composed of normal and anomal data.
The classes of MVTec dataset can be divided into two main categories depending on whether the align fits or not.

In the images of the left two columns in Figure \ref{fig:fig4}, align of first row and second row images does not fit.
In common, they do not align with rotation, therefore if first row images are rotated properly, they can be aligned with the second row images.
There are four classes with these characteristics in MVTec dataset : screw, hazelnut, metal nut, and grid.
These are denoted by \emph{Misaligned Classes of Original MVTec} in this paper.

On the other hand, in the images of the right two columns in Figure \ref{fig:fig4}, align of first and second row images fit.
Cable and Capsule images aligned well with rotation angle.
Leather and tile images are all composed of texture so they does not have any 2D alignment factors \eg, translation, rotation, scale, affine, perspective.
The classes with these characteristics are 11 classes, excluding the four classes of Misaligned Classes of Original MVTec.
These 11 classes are denoted by \emph{Aligned Classes of Original MVTec} in this paper.

\subsubsection{Synthetic MVTec w/ Alignment}
\label{subsec:+align mvtec}
Misaligned Classes of Original MVTec images can be converted to aligned well by each class if they are rotated appropriately.
Therefore, if we set a random normal image as template image for each class and rotate all images to align this image, we can convert MVTec Not Aligned to be aligned.
To this end, the self homography learning framework described in Section \ref{subsec:input foreground alignment} was used.
Original self homography learning is a method of learning the transformation applied to the aligned images.
By modifying this structure, the transformed image and the original image are concatenated and received as input.
At this time, since only rotation is the transformation to be learned, the ground truth was produced by converting rotation angle into 4 corners perturbation, not random perturbation.
The model that received the concatenated image was trained to regress the perturbation in the same way as the self homography learning method.
The trained model can match the angle of rotation to align two images of various forms.
Therefore, by using the trained model, we can rotate all images to be aligned by each class.
At this time, an empty space occurs when rotating the image, and we filled this space by reflection of original image.
This dataset is denoted by \emph{Synthetic MVTec w/ Alignment} in this work, and samples are in Figure \ref{fig:fig5}.
We can see that first and second row images in Figure \ref{fig:fig5} aligned well compared to Figure \ref{fig:fig4}.

\begin{table*}[tp!]
\setlength{\tabcolsep}{8pt}
\renewcommand{\arraystretch}{1.1}
\caption{{\ej Performance comparison of the state-of-the art studies between the pre-trained anomaly detection (PAD) and PAD with homography learning (HL) for investigating the effect of the degree of alignment.}}
\label{tbl:tbl2}
\begin{center}
\resizebox{\textwidth}{!}{%
\begin{tabular}{l|c|ccc|ccc}
\toprule
\multirow{2}{*}{\textbf{Method}} & \multirow{2}{*}{\textbf{Approach}} & \multicolumn{3}{c|}{\textbf{Image-level AUROC}} & \multicolumn{3}{c}{\textbf{Pixel-level AUROC}} \\ \cline{3-8} 
                                &                                     & Object & Texture & Total & Object & Texture & Total \\ 
\toprule
\multirow{2}{*}{PaDiM \cite{defard2021padim}}     
  & PAD        & 93.44 & 98.70 & 94.84 & 97.48 & 96.39 & 97.19 \\
  & PAD w/ HL  & 96.38 (\textbf{+2.94}) & 97.30 (-1.40) & 95.46 (\textbf{+0.62}) & 98.00 (\textbf{+0.52}) & 96.24 (-0.15) & 97.60 (\textbf{+0.41}) \\ \hline
\multirow{2}{*}{PatchCore \cite{roth2022towards}} 
  & PAD        & 97.97 & 99.23 & 98.31 & 97.59 & 96.29 & 97.24 \\
  & PAD w/ HL  & 98.28 (\textbf{+0.31}) & 94.87 (-4.35) & 97.29 (-1.02) & 97.75 (\textbf{+0.16}) & 94.69 (-1.60) & 97.08 (-0.17) \\ \hline
\multirow{2}{*}{SPADE \cite{cohen2020sub}}     
  & PAD        & 83.19 & 95.32 & 86.42 & 94.88 & 95.53 & 95.05 \\
  & PAD w/ HL  & 84.92 (+1.74) & 62.67 (-32.66) & 78.64 (-7.78) & 95.63 (\textbf{+0.75}) & 95.29 (-0.24) & 95.10 (\textbf{+0.05}) \\ \hline
\multirow{2}{*}{Mah.AD \cite{rippel2021modeling}}    
  & PAD        & 90.44 & 92.02 & 90.86 & - & - & - \\
  & PAD w/ HL  & 91.96 (\textbf{+1.52}) & 68.80 (-23.21) & 83.74 (-7.12) & - & - & - \\
\toprule
\end{tabular}%
}
\end{center}
\end{table*}

\subsubsection{Synthetic MVTec w/o Alignment}
As opposed to Section \ref{subsec:+align mvtec}, we converted MVTec dataset not to be aligned well.
To destroy align of MVTec dataset, we applied appropriate 2d transformation to MVTec images.
At this time, only rotation, translation, and scale were applied to preserve the original image form.
Perspective, affine transform are not applied because they can distort the original form.
Finally, the maximum transform was applied to the extent that the foreground was not cut
The dataset generated in this way is denoted by \emph{Synthetic MVTec w/o Alignment}.
Comparing the Synthetic MVTec w/o Alignment sample images of Figure \ref{fig:fig6} with the original MVTec dataset sample images in Figure \ref{fig:fig4}, it can be seen that the alignment is more distorted.

\subsection{Experimental Setup}
The PAD methods to be compared in this section are PaDiM \cite{defard2021padim}, Patchcore \cite{roth2022towards}, SPADE \cite{cohen2020sub}, and MahAD \cite{rippel2021modeling}.
Since MahAD cannot measure performance at pixel level, only image level AUROC was measured.

In Section \ref{subsec:effect of align}, We experimented on the effect of the align of the dataset on performance of PAD methods.
At this time, the AD performance for each PAD method was measured using Synthetic MVTec w/o Alignment, original MVTec dataset, Synthetic MVTec w Alignment.
Since the Synthetic MVTec w Alignment consists of only four classes: screw, hazelnut, metalnut, and grid, Synthetic MVTec w/o Alignment and original MVTec dataset were all experimented with only the same four classes for fair comparison.

In Section \ref{subsec:effect of SHL}, we experimented the effect of self homography learning on the performance of PAD.
The 11 classes of Aligned Classes of Original MVTec are fine tuned by self homography learning on original MVTec dataset.
But the 4 classes of Misaligned Classes of Original MVTec cannot be fine tuned by self homography learning because they do not aligned.
Therefore remaining 4 classes are fine tuned using Synthetic MVTec w/ Alignment.
The model was trained to regress homography matrix corresponding to the random perturbatiion of four corners to learn general 2D transformations.
At this time, if the four corners perturbation faces outside of the original image, there will be an empty space.
Therefore, to prevent this, the perturbation is directed inside the image.
Fine tuning performed 3000 iterations by each class and measured AUROC by each 100 iteration.
Among the performances of each iteration, the iteration with the highest average AUROC was selected as the representative value.
As for the augmentation in all experiments, both color augmentation such as hue and bright and shape augmentation such as pepper and salt were applied in common, and WideResNet50 \cite{zagoruyko2016wide} is used as the backbone network.

\subsection{Results}
\subsubsection{Effects of Alignment on Anomaly Detection}
\label{subsec:effect of align}
Table \ref{tbl:tbl1} is the result of experiment with the effect of the align of dataset on the PAD methods.
Most of the image level AUROC was higher than that of same method on worse aligned dataset.

Only image level AUROC of PatchCore in Synthetic MVTec w/ Alignment was lower than that of MVTec dataset.
This seems to be due to the influence of the padding value filling the empty space when generate Synthetic MVTec w/ Alignment.
Unlike other methods, PatchCore compares test feature patch with all train feature patches.
Therefore, the same patch of foreground exists in padding, and PatchCore measures this part the same score as foreground.
As a result, the decline in image level AUROC for the Synthetic MVTec w/ Alignment of PatchCore should be interpreted as a side effect of padding rather than the side effect of align.

\subsubsection{Effects of Self Homography Learning on Anomaly Detection}
\label{subsec:effect of SHL}


\begin{figure}
    \centering
    \includegraphics[width=160pt]{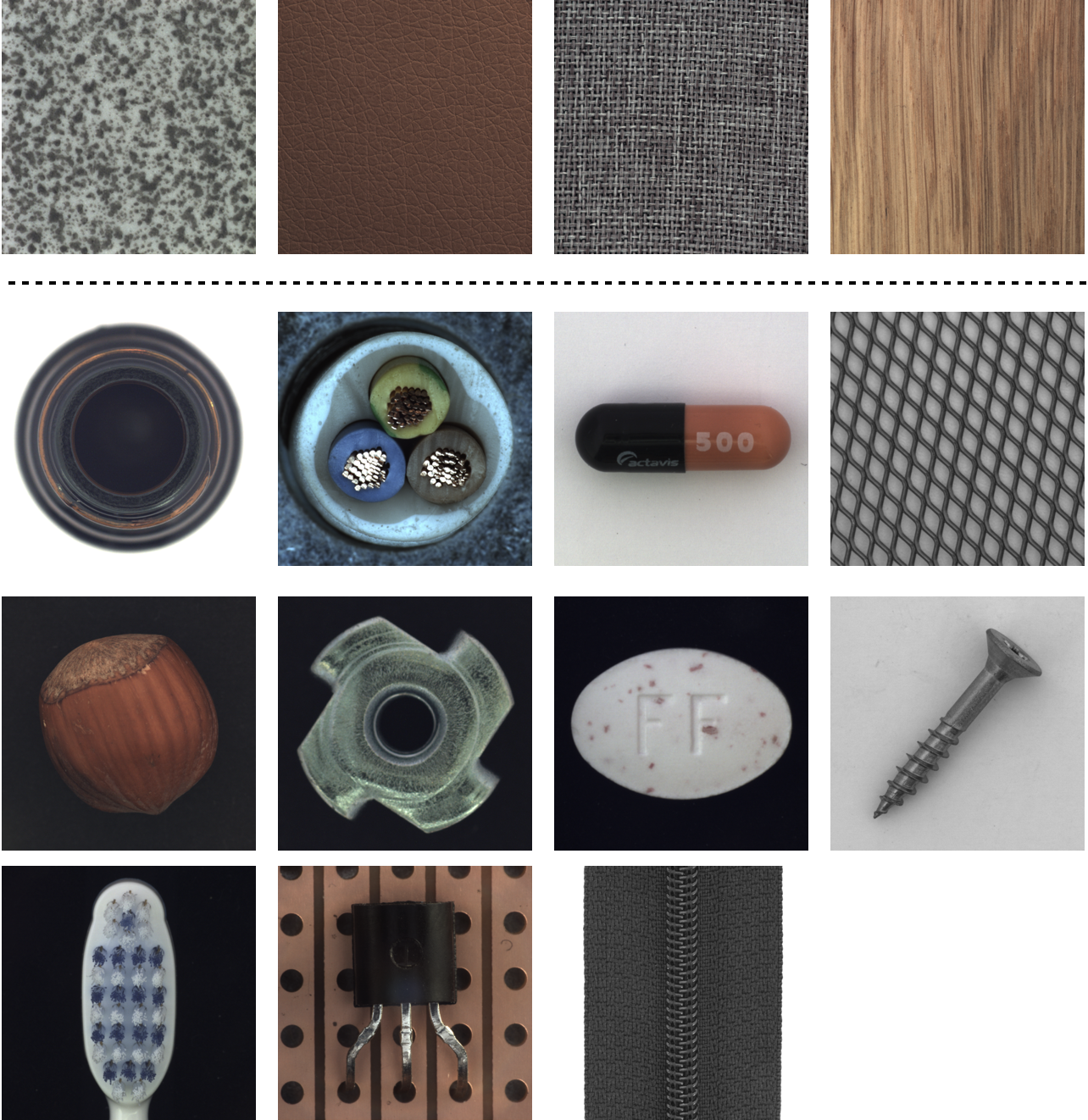}
    \caption{Image samples of texture classes (1st row) and object classes (2nd, 3rd, and 4th row). }
   \label{fig:fig7}
\end{figure}

Table \ref{tbl:tbl2} summarizes the experimental results to prove the effectiveness of our proposed self homography learning method.
Each row refers to an average AUROC using ImageNet pretrained network for PAD methods and an average AUROC using a fine tuned network by self homography learning.

We classified the performance into object classes' and texture classes'.
The texture class refers to a class with little shape information.
First row in Figure \ref{fig:fig7} is a sample images of texture classes.
They are characterized by a flat image when the input image is resized small for annomaly detection.
The texture classes we defined by these criteria are tile, leather, carpet, and wood.

On the other hand, the object class refers to a class composed of shape-oriented characteristics that define the class.
Images in 2nd, 3rd, 4th row of Figure \ref{fig:fig7} do not become flat images even if the input image is resized small, and shape information remains.

Interestingly, all PAD methods using self homography fine tuned network perform better on object classes then the ones using ImageNet pretrained network.
On the other hand, the performance on texture classes showed an opposite tendency.
The cause of these results will be analyzed later in analysis (See Section \ref{subsec:heatmap analysis}).

\section{Analysis}
In this section, the experimental results were analyzed.
In Section \ref{subsec:augmentation}, the performance change according to the combination of augmentation was analyzed using PaDiM.
In Section \ref{subsec:effect of backbone network}, we analyzed whether experimental results with the same tendency were obtained even when various backbones were used.
In Section \ref{subsec:heatmap analysis}, we used heatmap of PaDiM  to analyze why the effect of self homography learning is effective for object classes and ineffective for texture classes.

\subsection{Augmentation}
\label{subsec:augmentation}


\begin{table}[]
\caption{Performance comparison of PaDiM between the pre-trained anomaly detection (PAD) and PAD with homography learning (HL) for investigating the effect of augmentation. The two values in parentheses represent image level AUROC and pixel level AUROC, respectively.}
\label{tbl:tbl3}
\centering
\resizebox{\columnwidth}{!}{%

\begin{tabular}{l|c|ccc|c}

\toprule

\multirow{2}{*}{\textbf{Class}} & 
\multirow{2}{*}{\textbf{PAD}} & 
\multicolumn{3}{c|}{\textbf{PAD w/ HL}} &
\multicolumn{1}{c}{\multirow{2}{*}{\textbf{Max}}}                

\\ \cline{3-5}  & 
                          & Shape + Color & Shape        
                          & Color \\
\toprule
                          
Bottle  & \multicolumn{1}{c|}{(99.9,98.1)}  & (\textbf{100.0},\textbf{98.4})    & (\textbf{100.0},\textbf{98.4}) & \multicolumn{1}{c|}{(99.8,98.0)} & (100.0,98.4) \\

Cable                    & \multicolumn{1}{c|}{ (89.0, 96.1)}          & (93.4, 97.3)          &           (\textbf{95.2}, \textbf{98.2}) & \multicolumn{1}{c|}{ (90.8, 96.9)} &  (95.2,98.2) \\

Capsule                    & \multicolumn{1}{c|}{ (89.9, 98.5)}          & (\textbf{93.2}, \textbf{98.7})          &           (92.2, 98.6) & \multicolumn{1}{c|}{ (86.8, 98.3)} &  (93.2,98.7) \\
Carpet          & \multicolumn{1}{c|}{(99.9, 98.9)}          & (\textbf{100.0}, \textbf{99.1})          &           (99.8, 99.0) & \multicolumn{1}{c|}{ (99.7, 98.8)} &  (100.0,99.1) \\
Grid                    & \multicolumn{1}{c|}{ (95.9, 97.4)}          & (\textbf{99.2}, \textbf{97.5})          &           (99.1, \textbf{97.5}) & \multicolumn{1}{c|}{ (98.3, 96.7)} &  (99.2,97.5) \\
Hazelnut                    & \multicolumn{1}{c|}{ (97.0, 98.5)}          & (99.3, 98.9)          &           (97.1, \textbf{99.0}) & \multicolumn{1}{c|}{ (\textbf{99.4}, 98.4)} &  (99.4,99.0) \\
Leather                    & \multicolumn{1}{c|}{ (\textbf{100.0}, 99.2)}          & (\textbf{100.0}, \textbf{99.4})          &           (98.9, 99.2) & \multicolumn{1}{c|}{ (98.5, 99.3)} &  (100.0,99.4) \\
Metal Nut          & \multicolumn{1}{c|}{(97.5, 96.4)}          & (97.7, 97.7)          &           (96.6, \textbf{97.8}) & \multicolumn{1}{c|}{ (\textbf{98.0}, 97.5)} &  (98.0,97.8) \\
Pill                    & \multicolumn{1}{c|}{ (87.3, 94.3)}          & (\textbf{99.0}, 97.4)          &           (97.0, \textbf{97.7}) & \multicolumn{1}{c|}{ (96.6, 97.2)} &  (99.0,97.7) \\
Screw                    & \multicolumn{1}{c|}{ (89.0, 99.1)}          & (\textbf{94.5}, \textbf{99.4})          &           (92.9, \textbf{99.4}) & \multicolumn{1}{c|}{ (89.5, 99.2)} &  (94.5,99.4) \\
Tile                    & \multicolumn{1}{c|}{ (96.5, 92.8)}          & (97.5, 94.5)          &           (\textbf{98.2}, 95.8) & \multicolumn{1}{c|}{ (95.5, \textbf{96.2})} &  (98.2,96.2) \\
Toothbrush                    & \multicolumn{1}{c|}{ (99.2, 98.7)}          & (\textbf{100.0}, \textbf{98.9})          &           (\textbf{100.0}, \textbf{98.9}) & \multicolumn{1}{c|}{ (\textbf{100.0}, \textbf{98.9})} &  (100.0,98.9) \\
Transistor                    & \multicolumn{1}{c|}{ (97.4, 96.9)}          & (\textbf{98.0}, 97.6)          &           (97.8, \textbf{98.0}) & \multicolumn{1}{c|}{ (95.7, 97.1)} &  (98.0,98.0) \\
Wood                    & \multicolumn{1}{c|}{ (98.4, 94.8)}          & (99.0, 94.9)          &           (97.3, 92.0) & \multicolumn{1}{c|}{ (\textbf{99.6}, \textbf{95.1})} &  (99.6,95.1) \\
Zipper          & \multicolumn{1}{c|}{(85.8, 98.4)}          & (95.5, 98.8)          &           (\textbf{95.6}, \textbf{98.9}) & \multicolumn{1}{c|}{ (93.0, 98.6)} &  (95.6,98.9) \\ 
\toprule
\textbf{Average}                    & \multicolumn{1}{c|}{ (94.8, 97.2)}          & (\textbf{97.7}, \textbf{97.9})          &           (97.2, \textbf{97.9}) & \multicolumn{1}{c|}{ (96.1, 97.8)} &  (98.0,98.2) \\
\toprule
\end{tabular}%
}
\end{table}

\begin{table}[]
\caption{Performance comparison of PaDiM between the pre-trained anomaly detection (PAD) and PAD with homography learning (HL) for investigating the effect of backbone networks.}
\label{tbl:tbl4}
\resizebox{\columnwidth}{!}{%
\begin{tabular}{c|c|cc|cc}

\toprule
\multirow{2}{*}{\begin{tabular}[c]{@{}c@{}}\textbf{Backbone}\\ \textbf{Network}\end{tabular}} & \multirow{2}{*}{\begin{tabular}[c]{@{}c@{}}\textbf{Class}\\ \textbf{Type}\end{tabular}} & \multicolumn{2}{c|}{\textbf{Image-level AUROC}}             & \multicolumn{2}{c}{\textbf{Pixel-level AUROC}}             \\ \cline{3-6} 
                                                                            &                                                                       & \multicolumn{1}{c|}{\textbf{PAD}} & \textbf{PAD w/ HL}     & \multicolumn{1}{c|}{\textbf{PAD}} & \textbf{PAD w/ HL }    \\ 
\toprule

\multirow{3}{*}{Resnet18 \cite{he2016deep}}                                                   & Object                                                                & \multicolumn{1}{c|}{88.86}   & 92.05 (\textbf{+3.19}) & \multicolumn{1}{c|}{96.91}   & 97.04 (\textbf{+0.14}) \\ \cline{2-6} 
                                                                            & Texture                                                               & \multicolumn{1}{c|}{98.81}   & 96.20 (-2.62) & \multicolumn{1}{c|}{95.83}   & 95.11 (-0.72) \\ \cline{2-6} 
                                                                            & Total                                                                 & \multicolumn{1}{c|}{91.52}   & 93.16 (\textbf{+1.64}) & \multicolumn{1}{c|}{96.62}   & 96.53 (-0.09) \\ 
\toprule

\multirow{3}{*}{WideResnet50 \cite{zagoruyko2016wide}}                                               & Object                                                                & \multicolumn{1}{c|}{93.44}   & 95.18 (\textbf{+1.74}) & \multicolumn{1}{c|}{97.48}   & 97.74 (0.26)  \\ \cline{2-6} 
                                                                            & Texture                                                               & \multicolumn{1}{c|}{98.70}   & 97.44 (-1.27) & \multicolumn{1}{c|}{96.39}   & 96.25 (-0.14) \\ \cline{2-6} 
                                                                            & Total                                                                 & \multicolumn{1}{c|}{94.84}   & 95.78 (\textbf{+0.94}) & \multicolumn{1}{c|}{97.19}   & 97.34 (\textbf{+0.15}) \\ \toprule
\multirow{3}{*}{EfficientNet\_B5 \cite{tan2019efficientnet}}                                           & Object                                                                & \multicolumn{1}{c|}{97.13}   & 97.33 (\textbf{+0.20}) & \multicolumn{1}{c|}{94.90}   & 95.57 (\textbf{+0.67}) \\ \cline{2-6} 
                                                                            & Texture                                                               & \multicolumn{1}{c|}{99.43}   & 99.07 (-0.36) & \multicolumn{1}{c|}{92.13}   & 93.36 (\textbf{+1.23}) \\ \cline{2-6} 
                                                                            & Total                                                                 & \multicolumn{1}{c|}{97.75}   & 97.79 (\textbf{+0.05}) & \multicolumn{1}{c|}{94.16}   & 94.98 (\textbf{+0.82}) \\ \toprule
\end{tabular}%
}
\end{table}

\begin{figure*}[tp!]
     \centering
     \begin{subfigure}[b]{0.95\textwidth}
         \centering
         \includegraphics[width=\textwidth]{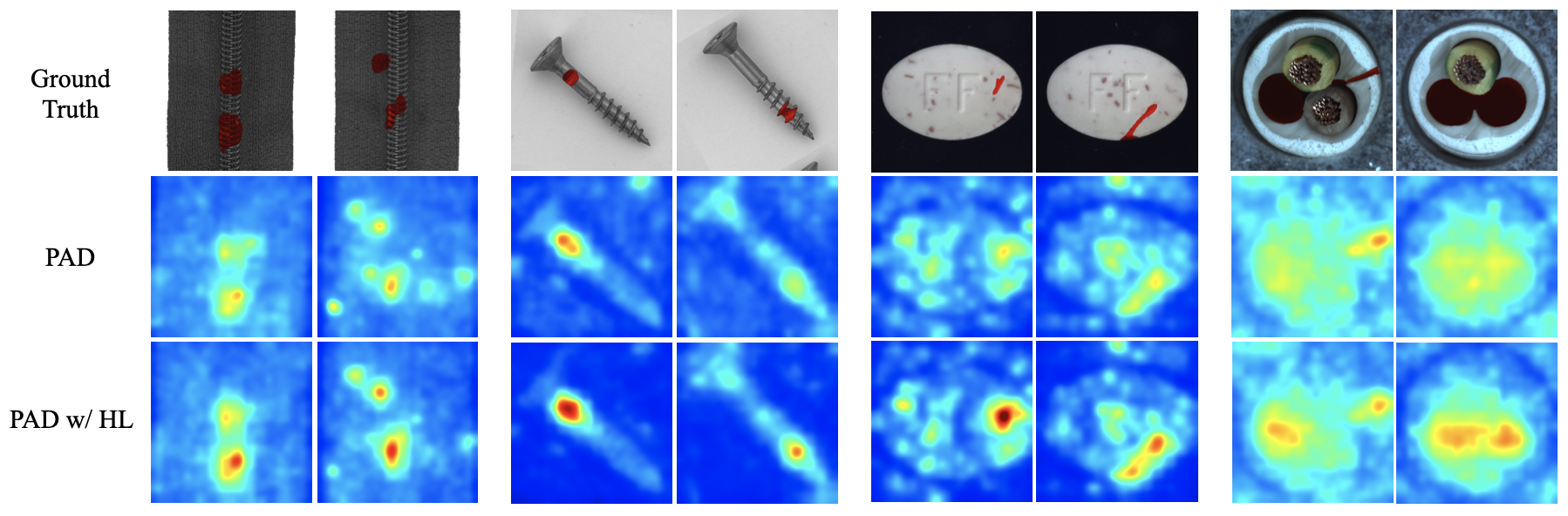}
         \caption{Heatmap for object classes, \eg, zipper, screw, pill, cable classes in  order from the left.}
         \label{fig:fig heatmap a}
     \end{subfigure}
     \begin{subfigure}[b]{0.95\textwidth}
        \label{fig:fig heatmap b}
         \centering
         \includegraphics[width=\textwidth]{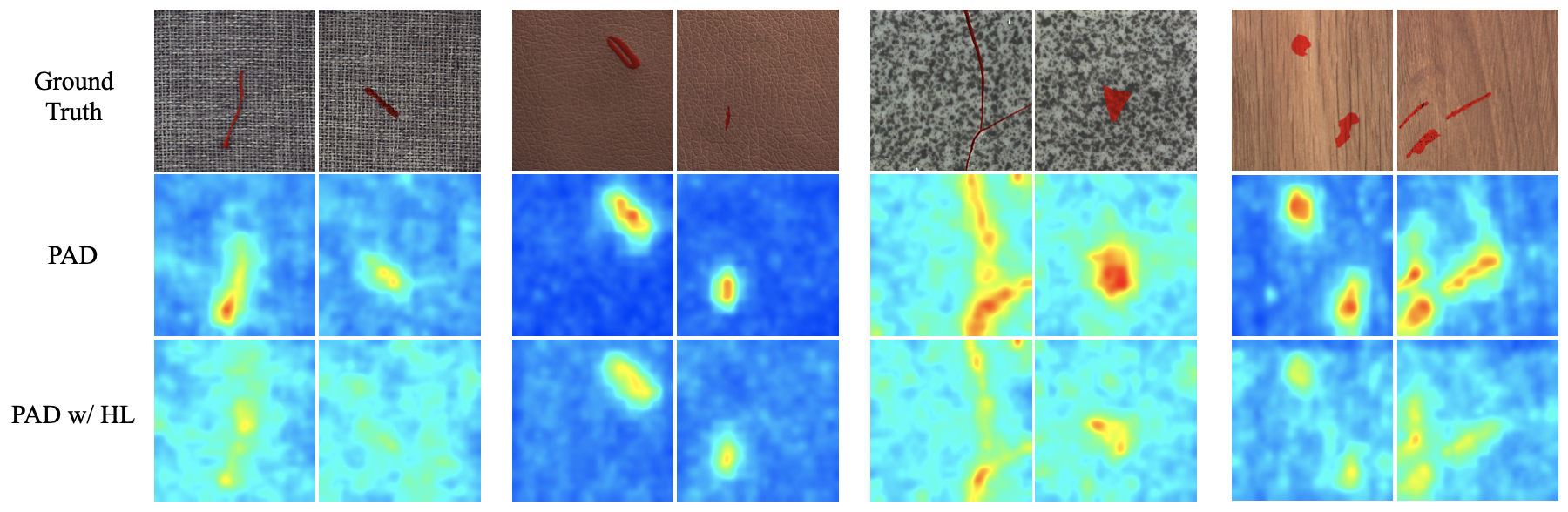}
         \caption{Heatmap for texture classes, \eg, carpet, leather, tile, wood classes in order from the left.}
         \label{fig:fig heatmap b}
     \end{subfigure}
        \caption{{\ej Visualization of the heatmap for the MVTec dataset. We compared heatmaps from the pre-trained anomaly detection (PAD) approach and PAD w/ homography learning (HL) with the ground truth images.}}
         \label{fig:fig8}
\end{figure*}

The final AD performance using self homography fine tuned network was different depending on which combination of augmentation was applied.
Table \ref{tbl:tbl3} summarizes the image and pixel level AUROC for each class according to the combination of augmentation.
The types of augmentation used in the experiment can be classified into shape and color.
The color category refers to the combination using hue and bright augmentation, and the shape category refers to the combination using pepper and salt augmentation.

As shown in Table \ref{tbl:tbl3}, There is at least one augmentation combination that perform better than using ImageNet pretrained network.
This means that when applying the self homography learning method in an actual industrial environment, even classes with a lot of textures can find better network than ImageNet pretrained network if using an appropriate combination of augmentation.

\subsection{Effect of Backbone Network}
\label{subsec:effect of backbone network}
We experimented whether the effect of self-homography learning has the same tendency in various backbone networks.
Table \ref{tbl:tbl4} summarizes the AUROC of MVTec dataset with PaDiM of various backbone networks.
Same As in Table \ref{tbl:tbl2}, the performance of the object classes increases and the performance of the texture classes decreases for all backbone networks.
These results imply that our proposed method has the same effect on various backbone networks.

\subsection{Heatmap Analysis}
\label{subsec:heatmap analysis}

Figure \ref{fig:fig8} summarizes the heatmap of the PaDiM using ImageNet pretrained network and the heatmap using self homography learning fine tuned network.

Figure \ref{fig:fig heatmap a} has a heatmap of object classes.
The heat map of the ImageNet pretrained network in the second row shows high scores in places around the foreground.
In addition, score of anomal part is not clearly larger than that of normal part.
On the other hand, in the heatmap using the self homography learning fine tuned network in the third row, foreground is clearly distinguished.
In addition, score of anomal part is clearly distinguished from the normal part.
This can be interpreted as fine tuned by self homography learning, the model learned which part is foreground and what characteristics the shape of foreground has.

In Figure \ref{fig:fig heatmap b}, there is a heatmap of texture classes.
Contrary to result of object classes, in the heatmap using the self homography learning fine tuned network, anomal points cannot be clearly distinguished.
This can be interpreted as, unlike the object class, the texture class has little shape information for learning homography, so the characteristics of the class were not well learned by self homography learning.


\section{Conclusion}
{\ej In this paper, we have experimentally demonstrated that the existing anomaly detection methods showed significant performance declines on datasets that do not align. In order to solve this problem, we proposed a novel deep anomaly detection framework based on the pre-trained network using a deep homography estimation method for real industrial environment, where the alignment is largely distorted. We found that just matching the alignment of the dataset significantly improved the performance of existing methods. Furthermore, we showed that the pre-trained network-based anomaly detection (PAD) approaches can be further improved by fine-tuning the ImageNet pre-trained network by self homography learning from normal data.

However, our framework has the following limitations: when performing the fine-tuning, the performance of the texture-oriented classes decreased slightly. It can be inferred that this is due to the loss of texture information during self homography learning that focuses on learning shape information. Thus, we need an effective method to supplement texture information loss during self homography learning. In addition, it would be more practical to make an end-to-end model because it is cumbersome to learn the homography estimation model for creating an aligned dataset in a real-world environment and feature fine tuning. Therefore, we'll leave these limitations as our future work.

}

\end{document}